\documentclass[a4paper,twoside]{article}
\usepackage{epsfig}
\usepackage{subcaption}
\usepackage{calc}
\usepackage{amssymb}
\usepackage{amstext}
\usepackage{amsmath}
\usepackage{amsthm}
\usepackage{multicol}
\usepackage{pslatex}
\usepackage{apalike} 
\usepackage{enumitem}
\usepackage{booktabs}
\usepackage[bottom]{footmisc}

\usepackage[titlenumbered,ruled]{algorithm2e}
\usepackage{enumitem}
\usepackage{graphicx}
\usepackage{float}
\usepackage{pgfplots}

\usepackage{tikz}
\usetikzlibrary{automata}
\usetikzlibrary{arrows}
\usetikzlibrary{shapes}
\usetikzlibrary{decorations.pathmorphing}
\usepackage{url}

\def\ST#1{\mathrm{SubTree}(#1)}
\def\STS#1{\mathrm{SubTreeSet}(#1)}
\def\TK#1#2{{\mathrm{TreeKernel}(#1, #2)}}
\def\STZ#1{\mathrm{SubTreeSeries}(#1)}
\def\PS#1{\mathrm{PrefixSet}(#1)}
\def\PSZ#1{\mathrm{PrefixSeries}(#1)}

\usepackage{SCITEPRESS}     

\newtheorem{df}{Definition}
\newtheorem{ex}{Example}

\begin{document}

%

\title{Parallel Tree Kernel Computation
}


\author{\authorname{Souad Taouti\sup{1}, Hadda Cherroun\sup{1} and  Djelloul Ziadi\sup{2}}
\affiliation{\sup{1}LIM, Université UATL Laghouat, Algeria}
\affiliation{\sup{2}Groupe de Recherche Rouennais en Informatique Fondamentale, Université de Rouen Normandie, France}
\email{souad.taouti@lagh-univ.dz, h.cherroun@gmail.com ,Djelloul.Ziadi@univ-rouen.fr}
}


\abstract{Tree kernels are fundamental tools that have been leveraged in many applications, particularly those based on machine learning for Natural Language Processing tasks. 
 In this paper, we devise a parallel implementation of the sequential algorithm for the computation of some tree kernels of two finite sets of trees~\cite{Nadia2015}. Our comparison is narrowed on a  sequential implementation of SubTree kernel computation.  This latter  is mainly reduced to an intersection of  weighted tree automata.  Our  approach relies on the nature of the data parallelism source inherent in this computation  by deploying  the MapReduce paradigm.  
One of the key benefits of our approach is its versatility in being adaptable to a wide range of substructure tree kernel-based learning methods.  To evaluate the efficacy of our parallel approach, we conducted a series of experiments that compared it against the sequential version using a diverse set of synthetic tree language datasets that were manually crafted for our analysis. The reached results  clearly demonstrate  that the proposed parallel algorithm outperforms the sequential one in terms of latency.}

\onecolumn \maketitle \normalsize \setcounter{footnote}{0} \vfill

\section{\uppercase{Introduction}}
\label{sec:introduction}
Trees are basic data structures that are naturally used in real world applications to represent a wide range of objects in structured form, such as XML documents~\cite{maneth2008xml}, molecular structures in chemistry~\cite{gordon1975chemistry} and parse trees in natural language processing~\cite{shatnawi2012parsetree}.

In~\cite{haussler1999convolution}, Haussler provides a framework based on convolution kernels, which find the similarity between two structures by summing the similarity of their substructures. Many convolution kernels for trees are presented based on this principle and have been effectively used to a wide range of data types and applications.

Tree kernels, which were initially presented in~\cite{NIPS2001_bf424cb7,Collins2002NewRA}, as specific convolution kernels, have been shown to be interesting approaches for the modeling of many real world  applications. Mainly those related to  Natural Language Processing tasks, e.g. named entity recognition and relation extraction~\cite{nasar2021}, text syntactic-semantic similarity~\cite{alian2023syntactic}, detection of text plagiarism~\cite{thom2018combining}, topic-to-question generation~\cite{chali2015towards}, source code plagiarism detection~\cite{source-code-2017}, linguistic pattern-aware dependency which  captures chemical–protein interaction patterns within biomedical literature~\cite{warikoo2018lptk}.

Subtree (ST) kernel~\cite{Vishwanathan2002FastKF} and the subset tree (SST)~\cite{NIPS2001_bf424cb7} were the initially kernels introduced in the context of trees. The compared segments in ST Kernel, are subtrees, a node and its entire descendancy. While in the SST Kernel, the considered segments are subset-trees, a node and its partial descendancy.

The principle of tree kernels, as initially presented, is to compute the number of shared substructures (subtrees and subset trees) between two trees $t_1$ and $t_2$ with $m$ and $n$ nodes, respectively. It may be computed recursively as follows:

\begin{equation}
    K(t_1, t_2)= \displaystyle\sum_{(n_1, n_2)  \in N_{t_1} \times N_{t_2}} \Delta (n_1, n_2)
\end{equation}

where $N_{t_1}$ and $N_{t_2}$ are the number of nodes in $t_1$ and $t_2$ respectively and $\Delta (n_1, n_2)= \sum^{|S|} _{i=1} I_i(n_1).I_i(n_2)$ for some finite set of
subtrees $S = \{s_1, s_2, \dots \}$, and $I_i(n)$ is an indicator function
which is equal to 1 if the subtree is rooted at node $n$ and to
$0$ otherwise.

An approach for computing the tree kernels of two finite sets of trees was proposed in~\cite{DBLP:journals/corr/MignotSZ15}. That makes use of Rooted Weighted Tree Automata (RWTA) that are a class of weighted tree automata. Subtree, Subset tree, and Rooted tree kernels can be computed using a general intersection of RWTAs associated with the  two finite sets of trees and then the computation of weights on the resulting automaton.

In this paper, we narrow  our study to the parallel implementation using MapReduce paradigm of the construction of the RWTA from a finite set of trees (as this step is a common base for other trees kernels) and to the comparison with the sequential linear algorithm proposed in~\cite{WardiSequential2023} for SubTree kernel computation. The main motivation behind this proposal is that despite the  linear complexity of the sequential version, it remains complex when we consider  Machine Learning computational cost requirements. 

We begin by defining RWTA. Next, we present the sequential proposed algorithms. Then we provide our parallel implementation based on MapReduce programming model.

The rest of the paper is organized as follows: Section~\ref{sec:TKA} introduces Tree Kernels and Automata while presenting the sequential proposed algorithms. Section\ref{sec:PTK} presents the details of the parallel implementation of the tree kernel computation based on MapReduce. Some experimental results and evaluations are shown in Section~\ref{sec:Experiments}. Finally, the conclusion and perspectives are presented in Section~\ref{sec:conclusion}

\section{\uppercase{Tree Kernels and Automata}}
\label{sec:TKA}

Let $\Sigma$ be a graded alphabet, $t$ is a tree over $\Sigma$, defined initially as $t = f(t_1, \dots , t_k)$ where $k$ can be any integer, $f$ any symbol from $\Sigma_k$ and $t_1, \dots , t_k$ are any $k$ trees over $\Sigma$. The set of trees over $\Sigma$ is referred as $T_\Sigma$. A tree language over $\Sigma$ is a subset of $T_\Sigma$.

Let $\mathbb{M}= (M, +)$ be a monoid with identity is $0$, a formal tree series $\mathbb{P}$~\cite{Collins2002NewRA}~\cite{esik2002formal} over a set $S$ represents a mapping from $T_\Sigma$ to $S$, where its support is the set $\text{Support}(\mathbb{P}) = \{t \in T_\Sigma | (\mathbb{P}, t) \neq 0 \}$. Any formal tree series is consistent to a formal sum $\mathbb{P} = \sum_{t \in T_{\sum}} (\mathbb{P}, t)t$, which is in this case both associative and commutative.

Weighted tree automata can realize formal tree series. In this paper, we employ specific automata, and the weights just indicate the finality of states. As a result, the automata we use are a particular subclasses of weighted tree automata.

\subsection{Root Weighted Tree Automata}
\begin{df}
Let $\mathbb{M}= (M, +)$  be a commutative monoid. An $\mathbb{M}$-Root Weighted Tree Automaton ($\mathbb{M}$-RWTA) is a 4-tuple $(\Sigma, Q, \mu, \delta)$ with the following properties: 

\begin{itemize}
    \item $\Sigma = \bigcup_{k \in \mathbb{N}} \Sigma_k$: a graded alphabet,
    \item $Q$: a finite set of states,
    \item $\mu$: the root weight function, is a function from $Q$ to $M$,
    \item $\delta$: the transition set, is a subset of $Q \times \Sigma_k \times Q^k$.
\end{itemize}
\end{df}  

\noindent  A $\mathbb{M}$-RWTA is referred  as RWTA, if there is no ambiguity.

\noindent  The root weight function $\mu$ is extended to $2^Q \to M$ for each subset $S$ of $Q$ by $\mu(S) = \Sigma_{s \in S} \mu(s)$. The function $\mu$ is equivalent to the finite subset of $Q \times M$ defined for any couple $(q, m)$ in $Q \times M$ by $(q, m) \in \mu \iff \mu(q) = m.$

\noindent  For any subset $S$ of $Q$, the root weight function $\mu$ is expanded to $2^Q \to M$ by $\mu(S) = \Sigma_{s \in S} \mu(s)$. The function $\mu$

\noindent The transition set $\delta$ corresponds to the function in $\Sigma_{k} \times Q^{k} \rightarrow 2^{Q}$ defined for any symbol $f$ in $\Sigma_k$ and for any $k$-tuple $(q_1, \dots , q_k)$ in $Q^k$ by

\begin{center}
    $q \in \delta(f, q_1, . . . , q_k) \iff (q, f, q_1,.\dots, q_k) \in \delta$.
\end{center}

\noindent  The function $\delta$ is extended to $\Sigma_{k} \times (2^Q)^k \rightarrow 2^Q$ as follows: for any symbol $f$ in $\Sigma_{k}$, for any $k$-tuple $(Q_1, \dots , Q_k)$ of subsets of $Q$,

\begin{center}
    $\delta(f, Q_1, \dots , Q_k) = \bigcup_{(q_1,\dots,q_k)\in Q_1 \times \dots \times Q_k} \delta(f, q_1, \dots , q_k)$.
\end{center}

\noindent  Finally, the function $\Delta$ is the function from $T_\Sigma$ to $2^Q$ defined for any tree $t = f(t_1, \dots , t_k)$ in $T_{\Sigma}$ by

\begin{center}
$\Delta(t) = \delta(f, \Delta(t_1), \dots , \Delta(t_k))$.
\end{center}

\noindent A weight of a tree  $t$ in a $\mathbb{M}$-RWTA $A$ is $\mu(\Delta(t))$. The formal tree series realized by $A$ is the formal tree series over $M$ denoted by $\mathbb{P}_A$ and defined by $\mathbb{P}_A=\sum_{t\in T_\Sigma}\mu(\Delta(t))$, with $\mu(\emptyset) = 0$ where $0$ is the identity of $\mathbb{M}$.    

\begin{ex}
Let us consider the graded alphabet $\Sigma$ defined by $\Sigma_0 = \{a, b\}$, $\Sigma_1 = \{g, h\}$ and $\Sigma_2 = \{f\}$. Let $\mathbb{M} = (\mathbb{N}, +)$. The RWTA $A = (\Sigma, Q, \mu, \delta)$ defined by

\begin{itemize}
    \item $Q = \{1, 2, 3, 4, 5, 6\},$
    \item $\mu = \{(1, 1), (2, 4), (3, 3), (4, 1), (5, 2), (6, 3)\},$
    \item $\delta$ = \{$(1, a), (3, b), (2, h, 1), (4, g, 3), (5, f, 2, 3),$
    $ (6, f, 4, 5)$\},
\end{itemize}

\noindent This RWTA  is represented in Figure~\ref{fig:rwta_a}. It realizes the following tree series: $\mathbb{P}_A = 1a + 3b + 4h(a) + 1g(b) + 2g(h(a)) + 3f(g(h(a)), g(b))$

\begin{figure}[ht]
  \centering
  \captionsetup{justification=centering}
  \scalebox{1}{
  \begin{tikzpicture}[node distance=2.5cm,bend angle=30,transform shape,scale=1]
   \node[state] (q6) {\( 6 \)};
    \node[state, below left of=q6] (q5) {\( 5 \)};
    \node[state, below right of=q6] (q4) {\( 4 \)};
    \node[state, below left of=q5] (q2) {\( 2 \)};
    \node[state, below of=q2] (q1) {\( 1 \)};
    \node[state, below of=q4] (q3) {\( 3 \)};
    \draw (q6) ++(-1cm,0cm) edge[above,<-] node {\( 3 \)} (q6);
    \draw (q5) ++(-1cm,0cm) edge[above,<-] node {\( 2 \)} (q5);
    \draw (q2) ++(-1cm,0cm) edge[above,<-] node {\( 4 \)} (q2);
    \draw (q4) ++(1cm,0cm) edge[above,<-] node {\( 1 \)} (q4);
    \draw (q3) ++(1cm,0cm) edge[above,<-] node {\( 3 \)} (q3);
    \draw (q1) ++(-1cm,0cm) edge[above,<-] node {\( 1 \)} (q1);
    \draw (q1) ++(0cm,-1cm) node {\( a \)}  edge[->] (q1);
    \draw (q3) ++(0cm,-1cm) node {\( b \)}  edge[->] (q3);
    \path[->]
    (q1) edge[->, above left] node {\( h \)} (q2)
    (q3) edge[->, above right] node {\( g \)} (q4)
    (q2) edge[->, above left] node {\( g \)} (q5);
    \draw (q4) ++(-1cm,1.0cm)  edge[->] node[above right,pos=0] {\( f \)} (q6) edge[shorten >=0pt,] (q4) edge[shorten >=0pt,] (q5);
  \end{tikzpicture}
 }
 \caption{The RWTA \( A \)}
 \label{fig:rwta_a}
\end{figure}
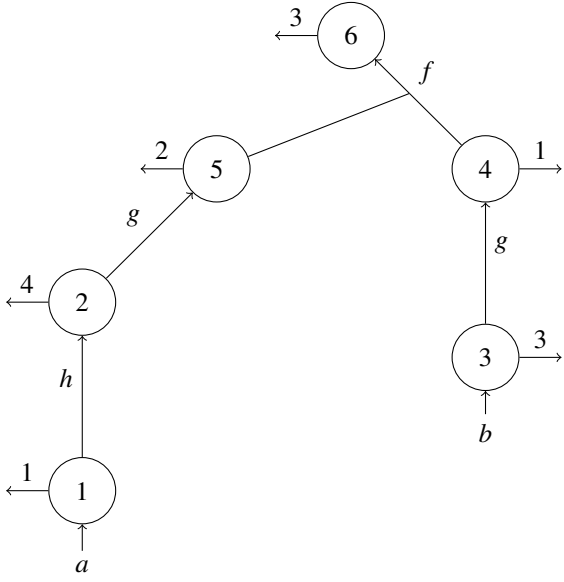

\end{ex}

A RWTA appears to be a prefix tree constructed in the context of words. A finite set of trees could be represented by this compact structure. 

In addition, many substructures' tree can be computed through this compact structure. In what follow, we introduce  Subtree,  Rooted tree and SubSet tree and their related automata~\cite{Nadia2015}.  However, in  this paper, we narrow our Kernel computation proposal to  the SubTree case. 
\noindent
\\

\subsection{Subtree}

\begin{df}
Let $\Sigma$ be a graded alphabet and $t = f(t_1, \dots , t_k)$ a tree in $T_\Sigma$. The set $\ST{t}$ is the set defined inductively by: 
\begin{center}
$\ST{t} = \{t\} \cup \bigcup_{1 \leq j \leq k} \ST{t_j}.$
\end{center}
\end{df}
\vspace{0.5em}

\noindent Let $L$ be a tree language over $\Sigma$. The set $\STS{L}$ is the set defined by:
\begin{center}
$\STS{L} = \bigcup_{t \in L} \ST{t}.$
\end{center}

\noindent The formal tree series $\STZ{t}$ is the tree series over $\mathbb{N}$ inductively defined by: 

\begin{center}
$\STZ{t} = t + \sum_{1 \leq j \leq k} \STZ{t_j}.$
\end{center}

\noindent If $L$ is finite, the rational series $\STZ{L}$ is the tree series over $\mathbb{N}$ defined by:

\begin{center}
$\STZ{L} = \sum_{t \in L} \STZ{t}.$
\end{center}

\begin{ex}
Let $\Sigma$ be the graded alphabet defined by $\Sigma_{0} = \{a, b\}$, $\Sigma_{1} = \{g, h\}$ and $\Sigma_{2} = \{f\}$.

Let  $t$ be the tree  $f(h(a), g(b))$, we have:
\newline
$\ST{t} = \{a, b, h(a), g(b), f(h(a), g(b))\}$ \newline
$\STS{t} = t + h(a) + g(b) + a + b$
\end{ex}

\begin{df}
Let $\Sigma$ be a graded alphabet. Let $t$ be a tree in $T_\Sigma$. The Subtree automaton associated with $t$ is the RWTA $A_t = (\Sigma, Q, \mu, \delta)$ defined by:
\begin{itemize}
    \item $Q =\STS{t}$,
    \item $\forall s \in Q,\quad \mu(s) = (\STZ{t}, s)$,
    \item $\forall f \in \Sigma, \forall s_1, \dots , s_{k+1} \in Q, s_{k+1} \in \delta(f, s_1, \dots, s_k) \iff  s_{k+1} = f(s_1, \dots , s_k).$
\end{itemize}
\end{df}

This  Weighted Tree Automaton (RWTA) requires less storage space because its states are exactly  its subsets.

\begin{ex}
Given the tree defined by $t = f(f(h(a),b),g(b))$. The RWTA $A_t$  associated with the tree $t$ is represented in Figure \ref{fig:rwta_t}.

\begin{figure}[H]
  \centering
  \captionsetup{justification=centering}
  \scalebox{1}{
  \begin{tikzpicture}[node distance=2.5cm,bend angle=30,transform shape,scale=1]
   \node[state, rounded rectangle] (q6) {\( t \)};
    \node[state, below left of=q6, rounded rectangle] (q5) {\( f(h(a), b) \)};
    \node[state, below right of=q6, rounded rectangle] (q4) {\( g(b) \)};
    \node[state, below left of=q5, rounded rectangle] (q2) {\( h(a) \)};
    \node[state, below of=q2, rounded rectangle] (q1) {\( a \)};
    \node[state, below of=q4, rounded rectangle] (q3) {\( b \)};
    \draw (q6) ++(-1cm,0cm) edge[above,<-] node {\( 1 \)} (q6);
    \draw (q5) ++(-1.5cm,0cm) edge[above,<-] node {\( 1 \)} (q5);
    \draw (q2) ++(-1cm,0cm) edge[above,<-] node {\( 1 \)} (q2);
    \draw (q4) ++(1cm,0cm) edge[above,<-] node {\( 1 \)} (q4);
    \draw (q3) ++(1cm,0cm) edge[above,<-] node {\( 2 \)} (q3);
    \draw (q1) ++(-1cm,0cm) edge[above,<-] node {\( 1 \)} (q1);
    \draw (q1) ++(0cm,-1cm) node {\( a \)}  edge[->] (q1);
    \draw (q3) ++(0cm,-1cm) node {\( b \)}  edge[->] (q3);
    \path[->]
    (q3) edge[->, above right] node {\( g \)} (q4)
    (q1) edge[->, above left] node {\( h \)} (q2);
    \draw (q2) ++(1cm,1.0cm)  edge[->] node[above left,pos=0] {\( f \)} (q5) edge[shorten >=0pt,] (q2) edge[shorten >=0pt,] (q3);
    \draw (q4) ++(-1cm,1.0cm)  edge[->] node[above right,pos=0] {\( f \)} (q6) edge[shorten >=0pt,] (q4) edge[shorten >=0pt,] (q5);
  \end{tikzpicture}
 }
  \caption{The RWTA \( A_{t} \) associated with the tree  \( t=f(f(h(a),b),g(b)) \).}
  \label{fig:rwta_t}
\end{figure}
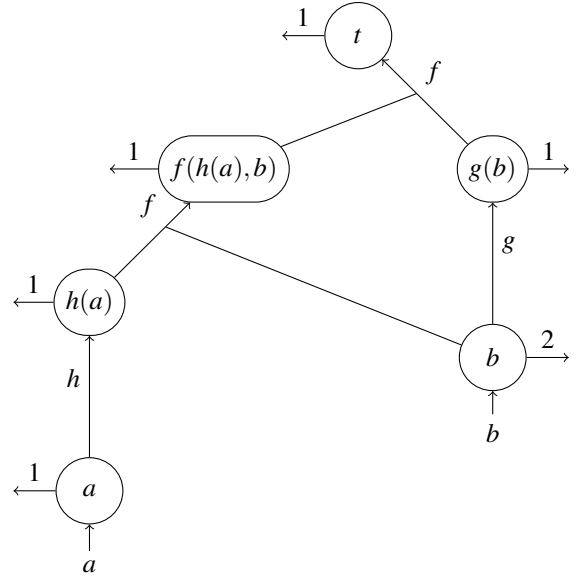

\end{ex}

\subsection{Rooted Tree}

\begin{df}
Let $\Sigma$ be a graded alphabet and $t = f(t_1, \dots , t_k)$ a tree in $T_\Sigma$. We indicate with $\Sigma^\prime$, the set $\Sigma \cup \{\perp\}$, where $\perp \in \Sigma ^\prime_0$ and $\perp \notin \Sigma$. 
$\PS{t}$ is the set of trees on $T_\Sigma^\prime$ inductively defined by:

$\begin{array}{ll}
 \PS{t} =     & \{t\} \cup f(\perp,\ldots, \perp) \\
        &\cup \ f(\PS{t_1},\ldots, \PS{t_k})
\end{array}$

\noindent It should be noted that $\perp$ is not a prefix of $t$.
\end{df}

\vspace{1em}
Let $L$ be a tree language over $\Sigma$. The set
PrefixSet($L$) is the set defined by:
\begin{center}
$\PS{L} = \bigcup_{t \in L} \PS{t}.$
\end{center}

The formal tree series $\PSZ{t}$ is the tree series over $\mathbb{N}$ inductively defined by: 

\begin{center} 
$\PSZ{t} = \sum_{t^\prime \in \PS{t}}  t^\prime$
\end{center}

\noindent If $L$ is finite, the series $\PSZ{L}$ is the tree series over $\mathbb{N}$ defined by:


$$\PSZ{L} = \sum_{t^\prime \in L}  \PSZ{t^\prime}$$

\noindent If $L$ is not finite, as $\Sigma$ is a finite set of symbols, there exists a symbol $f$ in $\Sigma^k$ such that $f(\perp, ..., \perp)$ occurs as a prefix infinite number of times in $L$. Therefore, $\PSZ{L}$ is a series of trees over $\mathbb{N} \cup \{+\infty\}$.

\begin{df}
Let $\Sigma$ be a graded alphabet. Let $t$ be a tree in $T_\Sigma$. The automaton of prefixes associated with $t$ is the RWTA $A_t = (\Sigma^\prime, Q, \mu, \delta)$ defined by:
\begin{itemize}
    \item $Q =\STS{t} \cup \{\perp\}$,
    \item $\forall t^\prime \in Q,\quad \mu(t^\prime) =$ 
    $ \begin{cases}
      1, & \text{if}\ t^\prime=t, \\
      0, & \text{else},
    \end{cases}$
    \item $\forall t^\prime = f(t_1,\dots, t_k) \in Q, \delta (f, t_1,\dots, t_k) = {t^\prime}$
    \item $\forall f \in \Sigma^k, \delta(f, \perp, \dots, \perp) = \{f(t_1,\dots, t_k)\in Q\}.$
\end{itemize}
\end{df}

\subsection{SubSet Tree (SST)}
\begin{df}
Let $\Sigma$ be a graded alphabet and $t = f(t_1, \dots , t_k)$ a tree in $T_\Sigma$. We indicate with $\Sigma^\prime$, the set $\Sigma \cup \{\perp\}$, where $\perp \in \Sigma ^\prime_0$ and $\perp \notin \Sigma$.
\newline
The set SSTSet($t$) is the set of trees on $T_{\Sigma^\prime}$ defined by:
$SSTSet(t) = PrefixSet(SubtreeSet(t))$
\end{df}
Let $L$ be a tree language over $\Sigma$. The set
SSTSeries($L$) is the set defined by:
\begin{center}
$SSTSeries(L) = \sum_{t^\prime \in L} SSTSeries(t) t^\prime.$
\end{center}
The formal tree series SSTSeries$(t)$ is the tree series over $\mathbb{N}$ inductively defined by: 
\begin{center}
$SSTSeries(t) = SubtreeSeries(prefixSet(t))$
\end{center}\begin{df}
Let $\Sigma$ be a graded alphabet. Let $t$ be a tree in $T_\Sigma$. The SST automaton associated with $t$ is the RWTA $A_t = (\Sigma^\prime, Q, \mu, \delta)$ defined by:
\begin{itemize}
    \item $Q =\STS{t} \cup \{\perp\}$,
    \item $\forall t^\prime \in Q,\quad \mu(t^\prime) =$ 
     $\begin{cases}
      1, & \text{if}\ t^\prime=t, \\
      0, & \text{else},
    \end{cases}$
    \item $\forall t^\prime = f(t_1,\dots, t_k) \in Q, \delta (f, t_1,\dots, t_k) = {t^\prime}$
    \item $\forall f \in \Sigma^k, \delta(f, \perp, \dots, \perp) = \{f_j(t_1,\dots, t_k)\in Q | h(f_j)=f\}.$
\end{itemize}
\end{df}

\subsection{Sequential Kernel Computation }

In order to compute the kernel of two finite tree languages $X$ and $Y$, we act in three steps: 
\begin{enumerate}
    \item First, we construct both RWTAs $A_{X}$ and $A_{Y}$.
    \item  Then we compute the intersection of $A_{X}$ and $A_{Y}$;
    \item Finally, the kernel is simply computed through a  sum of  all of the root weights of this RWTA.  
\end{enumerate} 

One can easily observe that  the set of states, denoted by  $Q$,  is equal to SubTreeSet$(t)$ in ST, plus $\{\perp\}$ for both Rooted Tree and SubSet Tree. However, their  root weight function ($\mu$ ) are different.
In addition, their  $\delta$,  denoting the transition functions,  are defined from the ST transition table. Consequently, the RWTA construction step remains  the same for the three tree substructures while the RWTA intersection step is distinct for each tree substructure. 

Let us recall that in this paper we narrow the RWTA intersection and kernel computation to the ST kernel case.

Initially, we introduce the sequential step-by-step procedure that enable us to efficiently calculate tree kernels by utilizing the intersection of tree automata. 

\subsubsection{RWTA Construction}

In this section, we describe through an algorithm the construction of an RWTA from a finite set of trees. Consider the finite set of trees $X$, first,  we extract all the prefixes of each tree in $X$ and sum their number of occurrences of each tree in $X$, which is equivalent to the sum of the subtrees series.
\begin{center}
    $\STZ{X} = \sum_{ t \in X} \STZ{t}$
\end{center}

Algorithm~\ref{alg:union} constructs an RWTA from a finite set of trees.

\begin{algorithm}
\SetKwInOut{Input}{Input}
\SetKwInOut{Output}{Output}
\Input{$X$: Set of trees}
\Output{RWTA $A_X$  = $(\Sigma, Q, \mu, \delta)$}  
$Q_{X} = \emptyset$\;
\ForEach{$t  \in X$}{

\uIf {$t \not\in Q_{X}$}{ 

Add($Q_X$, $t$);

$\mu_X(t) \gets 1$;
}

\Else{
    $\mu_X(s) \gets \mu_X(s) + 1$;
}
}
\caption{Computation of Automaton $A_X$ from $X$}
\label{alg:union}
\end{algorithm}

\subsubsection{RWTA Intersection}

\begin{df}
Let $\sum$ be an alphabet. Let $X$ and $Y$ be two finite tree languages over $\Sigma$. The Tree Series of  ($X$, $Y$) is defined by: 

$\STZ{(X, Y)}=$
\newline
$\sum_{t \in T_{\sum}}(\STZ{X}, t) \times (\STZ{Y}, t).$
\end{df}

\begin{ex}
Let $\Sigma$ be the graded alphabet defined by $\Sigma_0 = \{a, b\}$, $\Sigma_1 = \{g, h\}$ and $\Sigma_2 = \{f\}$. Let us consider the three  trees $t_1 = f(f(h(a), b), g(b))$, \ $t_2 = f(h(a), g(b))$ and $t_3 = f(f(h(a), b), f(h(a), g(b)))$. 

We have:
\newline

\noindent $\STZ{t{_1}} = t_1 + f(h(a), b) + h(a) + g(b) + a + 2b$
\newline

 \noindent $\STZ{t{_2}} = t_2 + h(a) + g(b) + a + b$
\newline

\noindent $\STZ{t{_3}} = t_3 + f(h(a), b) + t_2 + 2h(a) + g(b) + 2a + 2b$
\newline

\noindent $\STZ{\{t_1,t_2\}} = t_1 + t_2 + 2f(h(a), b) + 2h(a) + 2g(b) + 2a + 3b$
\newline

\noindent $\STZ{(\{t_1,t_2\}, \{t_3\})} = t_2 + f(h(a), b) + 4h(a) + 4g(b) + 4a + 3b$
\end{ex}

\vspace{0.5em}

\noindent By definition, any $t$ in $T_\Sigma$, is in $Q$ if and only if $t \in \STS{X} \cap \STS{Y}$. Moreover, by definition of $\mu$, for any tree $t$, $\mu(t) = \mu_X(t) \times \mu_Y(t)$, as for any tree $t$, if $t \notin \STS{X}$ (resp. $t \notin  \STS{Y})$, then $\mu_X(t) = 0$ (resp. $\mu_Y(t) = 0$).

In order to compute the kernel from two RWTA, Algorithm~\ref{alg:intersection} below, loops through the states of $A_X$, if any state $q$ is present in $A_Y$ then it is added to the automaton $A_{(X,Y)}$ with $\mu_{(X,Y)}(q) = \mu_X(q) \times \mu_Y(q)$, respectively to $A_Y$.

\begin{algorithm}
\SetKwInOut{Input}{Input}
\SetKwInOut{Output}{Output}
\Input{ $2$ RWTA  $A_X$ and $A_Y$}
\Output{an RWTA $A_{(X,Y)}=A_X \times A_Y$} 

$A_{(X, Y)} \gets  A_Y$;

\ForEach { $s \in Q_X$ }
{
\uIf {$s \in  Q_Y$}{ 
$\mu_{(X,Y)}(s) \gets \mu_X(s) \times \mu_Y(s)$;
}
\Else{
 $\mu(s) \gets 0$;
}
}
\caption{Computation of Automaton $ A_{(X,Y)} = A_X \times A_Y$}
\label{alg:intersection}
\end{algorithm}

\subsubsection{Subtree Kernel Computation}

After the construction of the RWTA for both sets of tree and then their intersection is built, in this last step we can effectively perform the  computation of tree kernel in subtree case. Let $X$ and $Y$ be two finite tree languages. Let $Z$ be the accessible part of their intersection tree automaton $A_{(X,Y)}$. Then, the kernel is simply computed through the sum of the weights:
\newline

$$\TK{X}{Y} = \sum_{q\in Z} \mu(q)$$.
\section{Parallel Tree Kernel Computation}
\label{sec:PTK}

According to the inherent data parallelism in the Tree Kernel Computation, we design and implement a parallel version of the previous sequential SubTree Kernel Computation  based on MapReduce paradigm. The framework provides parallel implementation in a distributed environment. It has some advanced features of distributed computing without programming to coordinate the tasks, as the large task is partitioned into small chunks that can be executed concurrently, which provides ease of handling job scheduling, fault tolerance, distributed aggregation, and other management tasks.

Kernel Computation are very  close to the Big Data paradigm which processing is very  challenging   compared with conventional data processing techniques. Numerous solutions have been properly proposed to address the computational and storage issues of big data. The  MapReduce framework is one of the prominent methods.

Before dealing with our parallel version, we explain the principle of MapReduce programming model in what follow. 

\subsection{MapReduce Framework}
Map-Reduce was created by Google as a parallel distributed programming approach that works on a cluster of computers due to large-scale data~\cite{Dayalan2004MapReduceSD}. It is termed parallel because tasks are executed by dedicating multiple processing units in a parallel environment, and distributed over distinct storage. Hadoop is among most popular  open-source MapReduce implementation created primarily by the Apache Software Foundation. \newline

\noindent MapReduce became the most popular framework for proposing programs without worrying about infrastructure complexity and difficulties since it provided such a stable, easy-to-use, abstract, and scalable environment. \newline
its  associated  programming paradigm is outlined through the basic Map-Reduce jobs.

\subsubsection{Map Function}
The first step of a Map-Reduce task is to map the input data to one or more specific reducers. Each mapper reads a set of files and generates a list of $<key, value>$ pairs for each unit of data based on a mapping schema. Key pairs from the same list would be collected in the same reducer. The most important element of a Map-Reduce method, which affects precision, time complexity, and space complexity, is the mapping schema.

\subsubsection{Reduce Function}
The output of mappers acts as input for the reducer function. Each reducer consequently receives a key associated with a list of records, denoted as $<key_{i}, record_{i_{1}}, record_{i_{2}},..., record_{i_{r}}>>$. Reducer output is also pairs of $<key, value>$, which can produce many values with the same keys. Within each reducer, the reducer function is applied in parallel to the input list of data and its output will be written to the Distributed File System.

\vspace{1em}
For our parallel   Subtrees Kernel Computation of two finite sets of trees $X$ and $Y$,  we act as in the sequential version in a  three-step process,  where two steps are performed  in parallel: 

\begin{enumerate}
    \item Parallel RWTA Construction,  
    \item Parallel RWTA Intersection,
    \item Kernel Computation which is deduced   by means of the same manner as in the sequential version. 
\end{enumerate}

\subsection{Parallel RWTA Construction}

Let $X$ be a finite tree language. To construct the RWTA $A_X$ from $X$ based on MapReduce paradigm, we have to  determine the Map and Reduce jobs.  The prefixes of $X$ are listed in a file which is  used as an input.

First, the Map function (Algorithm \ref{alg:map-union}) splits the prefixes list into subtrees. Next, the Map function distributes the key-value pairs as follows: $<key, 1>$, where the key represents the subtree, and the value is $1$ indicating the number of occurrence. At this step it represents its presence.

Then, the Map function sends the key-value pairs to  reducers by key, so if a subtree appears multiple times, it will be sent to the same reducer multiple times. This principle is similar to the popular word count program where words are subtree.

\begin{algorithm}
\SetKwInOut{Input}{Input}
\SetKwInOut{Output}{Output}
\Input{($X$: Set of trees) file}

\While{$file \neq empty$}{

Split$(\text{line}, S)$;

\ForEach{$s \in S$}{

Emit$(s, 1)$;
}

}


\caption{Map Function for the construction of $A_X$ from $X$}
\label{alg:map-union}
\end{algorithm}

Next, the reducer (Algorithm~\ref{alg:reduce-union}) sums the value which is the number of occurrences and is equivalent to the weight of the subree. Finally, the subtrees are merged in tree series.

\begin{algorithm}
\SetKwInOut{Input}{Input}
\SetKwInOut{Output}{Output}

\Input{mapped $\textlangle s, 1\textrangle$}

\Output{RWTA $A_X$}

Weight$_s \gets 0$;

\ForAll{all mapped $s$}{

$\text{Weight}_s \gets \text{Weight}_s + 1$;

}

Add$(A_X, (s,\text{Weight}_s))$;

\caption{Reduce Function for the construction of $A_X$ from $X$}
\label{alg:reduce-union}
\end{algorithm}

\subsection{Parallel RWTA Intersection}

Let $X$ and $Y$ be two finite tree languages, $A_X$ and $A_Y$ are their respective RWTA. In order to compute the intersection of automata $A_X$ and $A_Y$ using MapReduce programming model, we have to identify the Map and Reduce jobs. Let us mention that we have both RWTA details saved, from the last Construction step,  as an input file, each of them in a separate line. 

First, the Map function (Algorithm~\ref{alg:map-intersection}) splits the tree series for $X$ and $Y$ resp. into subtrees with their weights in list. Then, the Map function distributes the key-value pairs as follows: $\textlangle \text{subtree}, (1, weight)\textrangle $, where the key represents the subtree, and the value is composed of the weight of the subtree and the number $1$ that acts as a Boolean indicating whether or not the subtree is present in the RWTA.

Next, the Map function distributes its output to the reducers by key i.e if any subtree is present in different RWTA, it will be sent to the same reducer. 

\begin{algorithm}
\SetKwInOut{Input}{Input}
\SetKwInOut{Output}{Output}
\Input{(RWTA $A_X$, $A_Y$) file}



\While{$file \neq empty$}{

Split$(\text{line}, Q)$;

\For{$s \in Q$}{

Emit$(s, (1, \mu(s)))$;
}}

\caption{Map Function for the computation of the automaton $A_X \times A_Y$}
\label{alg:map-intersection}
\end{algorithm}

After that, every reducer (Algorithm~\ref{alg:reduce-intersection}) sums the first part of the value, which indicates the presence of the subtree in the RWTA to check if it is present in both RWTAs. Then, if the presence is equal to 2 the reducer multiplies the weights of the received subtree from $A_X$ and $A_Y$.

\begin{algorithm}
\SetKwInOut{Input}{Input}
\SetKwInOut{Output}{Output}
\Input{$\textlangle s, (1, \mu(s)) \textrangle $}
\Output{RWTA $A_X \times A_Y$}

Presence$_s \gets 0$;

Weight$_s \gets 1$;

\ForAll{ mapped $s$}{

$\text{Presence}_s \gets \text{Presence}_s + 1$;

$\text{Weight}_s \gets \text{Weight}_s \times \mu(s)$; 
}

\If {Presence$_s = 2$}{ 

Add$(A_X \times A_Y, (s, \text{Weight}_s))$;
}
\caption{Reduce Function for the computation of the intersection  automaton $A_X \times A_Y$}
\label{alg:reduce-intersection}
\end{algorithm}

\section{\uppercase{Experiments and Results}}
\label{sec:Experiments}
To analyse our parallel RWTA-based SubTree Kernel computation we have performed a batch of  comparative experiments in order to demonstrate the difference in terms of latency   between our parallel  algorithm and the sequential  one using MapReduce paradigm. 

Additionally,  we use the absolute acceleration metric defined by $A^{abs} = T_{seq}/T_{par}$ where $T_{seq}$ and $T_{par}$ are the running times of the sequential  and the parallel algorithms respectively.

\noindent Prior to presenting our findings, let us initially provide a description of the benchmark we constructed and outline the implementation details.

\subsection{Dataset}

In order to perform the  comparative study  of both variants of algorithms, we need  a testbed of  multiple datasets that cover the  variety wide of tree characteristics in order to have a deep algorithm analysis, which is not the case in the real world  datasets that are standard benchmarks for learning on relatively small trees. For that purpose, in our experiments, we are brought in   generating  synthetic datasets. 

For this building  dataset task,  we have considered into account mainly three criteria:  
i) the  alphabet size (varying between $2$ and $12$), 
ii) the range of the  maximal alphabet arity (between $1$ and $5$), and   
iii) the tree depth ($TD$) that we have varied within the range of  $10$ to $50$.  

The constructed tree datasets are  divided  into two batches   according to the alphabet size in [$2$,$12$]. The first batch (respectively the second batch) gathered three  datasets      $D1$, $D2$, $D3$ and $D4$ (respectively $D1^s$, $D2^s$, $D3^s$ and $D4^s$ ). Each of them contains  two  tree sets generated using this above principle.  Into each batch, each dataset has different size that we have classified  according to their average tree size in three classes (small: less or equal to  $1.5$GB, medium: less than  $2.5$GB, and large: more than  $4.4$GB and $7$GB). 
Table~\ref{tab:datasets} illustrates more details on the generated datasets. 

\begin{table}[h]
    \centering
    \begin{tabular}{lrrrrr}
    \toprule
     &\#~Trees   &$\Sigma$ & Arity  & $TD$ &  size Gb \\\midrule
$D1$   & 500 &  [2, 12]   &  [1, 5]  &[10,20]   & $ 1.5$  \\\midrule
$D2$   & 800  & [2, 12]  &  [1, 5]  & [10, 50] & $ 2.5$   \\\midrule
$D3$   & 3000  & [2, 12]  &  [2, 5]  & [10, 50]  &  $ 4.4$
\\\midrule
$D4$   & 4500  & [2, 12]  &  [2, 5]  & [10, 50]  &  $ 7$ \\\midrule\midrule
$D1^s$   & 500 &  [2, 6]  &  [1, 5]  &[10,20]   & $ 1.5$  \\\midrule
$D2^s$   & 800  & [2, 6] &  [1, 5]  & [10, 50] & $ 2.5$ \\\midrule
$D3^s$   & 3000  & [2, 6]  &  [2, 5]  & [10, 50]  &  $ 4.4$  
\\\midrule
$D4^s$   & 4500  & [2, 6]  &  [2, 5]  & [10, 50]  &  $ 7$  
\\\bottomrule  
    \end{tabular}
    \caption{Details on the generated datasets.}
    \label{tab:datasets}
\end{table}

Both sequential and MapReduce-based algorithms are implemented in Java~11. All experiments were performed on a server equipped with an Intel(R) Xeon(R) Silver $4216$ CPU (2.10GHz) processor with  $32$ cores and $128$GB of RAM running Linux.

Sequential and parallel   codes  in addition to the  generated datasets are available on Github~\footnote{ \url{https://github.com/souadtaouti/parallel-Subtree-kernel-computation-using-RWTA/}}.

In this study, we have established a fixed cluster architecture, utilizing Docker containers for conducting all tests. Our cluster comprises one container designated as the Master, and five containers serving as Slaves. It is important to note that this choice of cluster configuration was made arbitrarily for the purpose of this research. 
 The Hadoop~\footnote{\url{https://hadoop.apache.org/}} V 3.3.0 is installed as MapReduce implementation platform on our cluster.

\subsection{Results}

\begin{figure}
    \centering
     \scalebox{0.9}{
\begin{tikzpicture}
  \begin{axis}[
 xbar,
    y axis line style = { opacity = 0 },
    axis x line       = none,
    tickwidth         = 0 pt,
    enlarge y limits  = 0.2,
    enlarge x limits  = 0.02,
    symbolic y coords = {$D4$,$D3$,$D2$,$D1$},
    nodes near coords,
  ]
  
 \addplot coordinates { (97.2,$D1$)   
 (139.7,$D2$) 
 (197,$D3$)
 (431,$D4$)};
 \addplot coordinates {(0.87,$D1$) (1.96,$D2$)  
 (3.75,$D3$)
 (8.67,$D4$)};
\legend{Sequential,Parallel}
\end{axis}

\end{tikzpicture}
}
 \caption{Performances of Parallel algorithm  vs the  Sequential one in terms of running time (minutes).}
    \label{fig:comparative}
\end{figure}
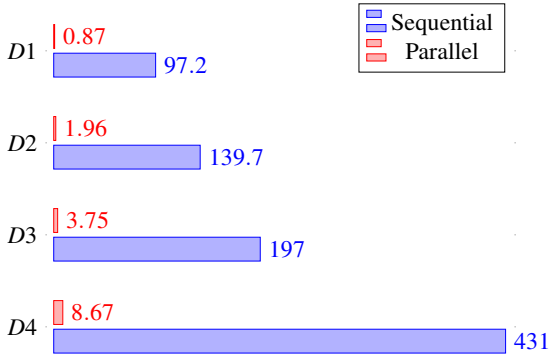

Figure~\ref{fig:comparative} reports the performances of  our parallel SubTree Kernel computation versus the sequential algorithm in terms of running time on the generated datasets. 
Let us mention that the sequential time is obtained on one node (container) of  our  cluster.

It is obvious to observe that  our parallel computation  is significantly faster compared to the sequential version across all dataset instances. Furthermore, our analysis reveals that the average absolute acceleration achieved is  
 $50,89$ times,  as demonstrated  in Table~\ref{tab:acceleration}.  This substantial acceleration is notable, which is a good acceleration, taking into account the utilization of six nodes in our cluster, and reflects the effectiveness of our parallel computation approach.

\begin{table}[!h]
    \centering
    \begin{tabular}{lrrrrrr}
    \toprule
                & $D1$      & $D2$  & $D3$  & $D4$ && \textbf{Average} \\\midrule
$A^{abs}$    & $57,23$    & $62.96$ & $42.61$ & $40.75$ &&$50,89$ \\\bottomrule 
    \end{tabular}
    \caption{Acceleration  of Parallel SubTree Kernel computation on different  datasets.}
    \label{tab:acceleration}
\end{table}

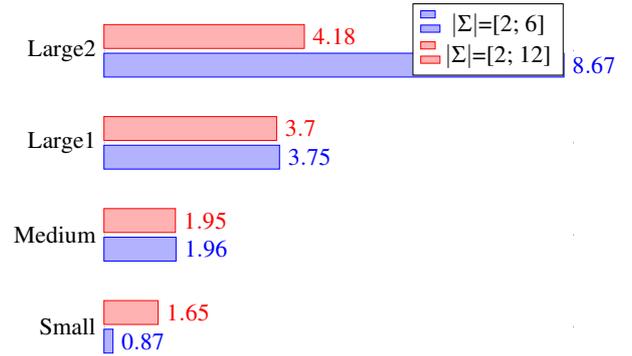
\begin{figure}[h]
   \scalebox{0.9}{
\begin{tikzpicture}
  \begin{axis}[
   xbar,
    y axis line style = { opacity = 0 },
    axis x line       = none,
    tickwidth         = 0pt,
    enlarge y limits  = 0.2,
    enlarge x limits  = 0.02,
    symbolic y coords = {Small,Medium,Large1,Large2},
   nodes near coords,
  ]
 \addplot coordinates { (0.87,Small) (1.96,Medium)  (3.75,Large1)  (8.67,Large2)};
\addplot coordinates {  (3.7,Large1) (1.95,Medium)  (1.65,Small)
(4.18,Large2)};
\legend{$|\Sigma|$=[2; 6], $|\Sigma|$=[2; 12]}
\end{axis}
\end{tikzpicture}
}
 \caption{Performances of our  Parallel algorithm in terms of running time (minutes)  when varying  the alphabet size.}
    \label{fig:comparative-sigma}
\end{figure}

Considering the influence of the  alphabet size variation on the performances of our algorithm,  we have measured  the performances reached by both datasets batches  (\{$D1$, $D2$,  $D3$ and and $D4$\}  and \{ $D1^s$, $D2^s$, $D3^s$  and $D4^s$\}. 
In Figure~\ref{fig:comparative-sigma}, we present the performance results of our parallel SubTree Kernel computation, showcasing the impact of the Alphabet size on the running time for the generated datasets.

The results demonstrate that our parallel algorithm performs better when the alphabet size is larger  at least for our datasets. This outcome is in line with our prediction, as a smaller alphabet size implies a higher frequency of common prefixes. A more detailed analysis of the number of common prefixes further supports this finding.

\section{\uppercase{Conclusion}}
\label{sec:conclusion}



The prefix tree automaton constitutes a common base for the computation of different tree kernels: SubTree, RootedTree, and SubSequenceTree kernels \cite{Nadia2015}. In this paper, we have shown a parallel Algorithm that efficiently compute this common structure (RWTA automaton) and we have used it for the computation of the SubTree Kernel using MapReduce paradigm. \\
Our parallel implementation of the SubTree kernel computation  has been tested on  synthetic datasets with different parameters. The results showed that our parallel computation is by far more speed than the sequential version for all instances of datasets.
Despite that this work has shown the efficiency of the parallel implementation compared to the sequential algorithms, three main future works  are envisaged.
Firstly, we have to devise some algorithms that generalise the computation of others kernels such  RootedTree, and SubSequenceTree \ldots. 
Some of them  will deploy tree automata intersection  in addition to the associated weights computation. In fact, while the subtree kernel is a simple summation of weights, the SubSequenceTree needs more investigation on the weight computations using  the resulted RWTAs intersection.   
Secondly, more large datasets have to be generated and tested to confirm the output-sensitive results of our solutions.
Finally, one can investigate different cluster architectures in order to give more insights and recommendations on the cluster' parameters tuning.


\bibliographystyle{apalike}
{\small \bibliography{conf}}

\end{document}